\documentclass{article} 
\usepackage{iclr2026_conference,times}

\usepackage{amsmath,amsfonts,bm}

\def\eqref#1{equation~\ref{#1}}

\def\1{\bm{1}}

\DeclareMathAlphabet{\mathsfit}{\encodingdefault}{\sfdefault}{m}{sl}
\SetMathAlphabet{\mathsfit}{bold}{\encodingdefault}{\sfdefault}{bx}{n}

\usepackage{url}
\usepackage{booktabs}
\usepackage{multirow}
\usepackage{array}
\usepackage{makecell}
\usepackage{xspace}
\usepackage[justification=raggedright,singlelinecheck=false]{caption}
\usepackage{tabularx}
\usepackage{graphicx}
\usepackage{subcaption}
\usepackage{caption}
\usepackage[ruled,vlined]{algorithm2e}
\usepackage{wrapfig}

\usepackage[colorlinks=true, citecolor=blue, linkcolor=blue, urlcolor=blue]{hyperref}

\newcommand{\ours}{\textsc{ReXMoE}\xspace}
\newcommand{\oursshort}{\textsc{ReX}\xspace}

\title{\ours: Reusing Experts with Minimal \\Overhead in Mixture-of-Experts}

\author{\bf
    Zheyue Tan$^{1,}$\thanks{Work conducted during internship at Infinigence-AI.} \quad Zhiyuan Li$^2$ \quad Tao Yuan$^2$ \quad Dong Zhou$^2$ \quad Weilin Liu$^2$ \quad Yueqing Zhuang$^2$  \\ 
    \bf
    Yadong Li$^2$ \quad Guowei Niu$^2$ \quad Cheng Qin$^3$ \quad Zhuyu Yao$^2$ \quad Congyi Liu$^2$ \quad Haiyang Xu$^2$ \\
    \bf
    Boxun Li$^2$ \quad Guohao Dai$^{4,5,}$\thanks{Corresponding authors.} \quad Bo Zhao$^{1,\dagger}$ \quad Yu Wang$^{6,\dagger}$ \\
    \normalsize \textsuperscript{1} Aalto University \quad
    \normalsize \textsuperscript{2} Infinigence-AI \quad
    \normalsize \textsuperscript{3} Yale University \quad
    \normalsize \textsuperscript{4} Shanghai Jiao Tong University \\
    \normalsize \textsuperscript{5} Shanghai Innovation Institute \quad
    \normalsize \textsuperscript{6} Tsinghua University
}

\iclrfinalcopy 
\begin{document}

\maketitle

\begin{abstract}

    Mixture-of-Experts (MoE) architectures have emerged as a promising approach to scale Large Language Models (LLMs). 
    MoE boosts the efficiency by activating a subset of experts per token.
    Recent works show that \emph{fine-grained experts} substantially enriches the combinatorial flexibility of active experts and enhances model expressiveness.
    However, such a design is fundamentally limited by the \emph{layer-local} routing mechanism: each layer is restricted to its own expert pool. 
    This requires a careful trade-off between expert dimensionality and routing diversity given fixed parameter budgets.
    We describe~\ours, a novel MoE architecture that improves routing beyond the existing layer-local approaches by allowing routers to \underline{re}use e\underline{x}perts across adjacent layers. 
    \ours decouples expert dimensionality from per-layer budgets, enabling richer expert combinations without sacrificing individual expert capacity or inflating overall parameters. 
    To this end, we propose a new \emph{progressive scaling routing} (PSR) strategy to gradually increase the candidate expert pool during training. 
    As a result, \ours improves both language modeling and downstream task performance.
    Extensive experiments on models ranging from 0.5B to 7B parameters across different architectures demonstrate that~\ours consistently improves performance under fixed architectural dimensions,
    confirming~\ours as new design paradigm for parameter-efficient and scalable MoE-based LLMs.

\end{abstract}

\section{introduction}

Large Language Models (LLMs) have rapidly advanced in scale and capability, reaching hundreds of billions of parameters and demonstrating remarkable progress toward Artificial General Intelligence (AGI). 
Recent foundation models~\citep{achiam2023gpt,openai_gpto3_o4,meta2025llama4,guo2025deepseek,kimi-k2,Yang2025Qwen3TR} have exhibited strong performance across complex tasks in multiple domains. 
This progress has been driven by massive investments in data and compute, but such growth also intensifies the tension between model capacity and development practicality. 
Given the substantial costs involved, Mixture-of-Experts (MoE) architectures have become an increasingly attractive alternative. 
By dynamically activating only a subset of specialized experts per input, MoEs can match or even exceed the performance of dense counterparts while significantly reducing inference-time computational demands~\citep{shazeer2017outrageously,lepikhin2020gshard,fedus2022switch,du2022glam,jiang2024mixtral,liu2024deepseek,liu2024deepseekv3,Yang2025Qwen3TR}.

Comparing to the dense counterparts, a key characteristic of MoE architectures is the additional degrees of freedom when replacing the feed-forward networks with MoE blocks: the number of experts, the dimensionality of each expert, and the routing strategy. 
Recent studies on MoE scaling laws~\citep{clark2022unified_moe_scaling_law,krajewski2024scaling_moe_law} highlight that model performance is constrained by trade-offs among these dimensions under a fixed parameter budget.
In particular, the size of each expert and the number of experts form a critical axis: increasing the number of smaller experts enriches the space of expert combinations, whereas larger experts preserve stronger representational capacity but limit routing diversity. 
Such a trade-off is the core of the MoE architectural design.

In practice, recent works show trends toward adopting \emph{finer-grained experts} in MoE design.
For example, early Mixtral-of-Experts models~\citep{jiang2024mixtral} employed 8 candidate experts per layer, whereas more recent models such as Qwen3~\citep{Yang2025Qwen3TR} series expand this to 128 experts, DeepSeek-V3~\citep{liu2024deepseekv3} scales the design to 256 experts. 
From a combinatorial perspective, fragmenting experts into smaller units substantially increases the number and diversity of possible routing combinations, thereby enhancing the expressiveness of MoE models and improving their ability to capture more specialized knowledge~\citep{dai2024deepseekmoe}.

A key challenge in existing MoE designs lies in the \emph{layer-local} routing mechanism, where each layer’s router is restricted to its own expert pool.
This constraint ties architectural choices to per-layer budgets and prevents more flexible balancing between the capacity of individual experts and the combinatorial flexibility of the expert pool.
As a result, finer granularity comes at the cost of reduced representational capacity for each expert, since smaller experts correspond to reduced hidden dimensionality in their feed-forward networks. 
On the other hand, preserving expert dimensionality while simply increasing the number of experts inflates the overall parameter count. 
This fundamental challenge motivates us to explore new architectural directions that enrich expert combinations without reducing expert capacity or inflating model size.

\begin{figure}[t]
    \setlength{\belowcaptionskip}{-15pt}
    \centering
    \includegraphics[width=0.9\linewidth]{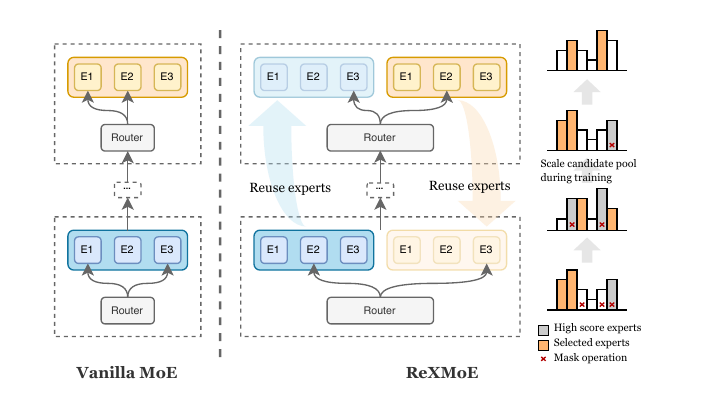}
    \caption{\textbf{Overview of~\ours.}
    Compared to vanilla MoE, \ours enables more flexible expert combinations by reusing experts from adjacent layers. 
    The only additional overhead comes from the router, which learns to route tokens to the expanded candidate pool. 
    During training, \ours progressively scale the candidate pool by gradually reducing the number of masked experts until all experts are available.
    }

    \label{fig:ours}
\end{figure}

To address such a challenge, we propose~\ours, a novel approach to MoE architecture design that extends routing beyond the conventional \emph{layer-local} boundary.
By allowing routers to reuse experts across grouped adjacent layers, \ours decouples expert dimensionality from per-layer parameter budgets and introduces a new dimension in MoE design: models can realize richer expert combinations without sacrificing individual expert capacity or inflating the total parameter count.
Furthermore, we present a Progressive Scaling Routing (PSR) strategy for training, which enhances the performance of models with reused expert pools.
As illustrated in~\autoref{fig:ours}, \ours reuses experts across adjacent layers with only negligible additional router parameters, while PSR gradually expands the candidate expert pool during training.
Extensive experiments on models ranging from 0.5B to 7B parameters across different architectures show that~\ours consistently improves performance under fixed dimensionality configurations.
In addition, ablation studies highlight key design factors and confirm the practicality of our approach.
Qualitative analysis further suggests that~\ours enhances task-specific specialization.
Together, these results establish~\ours as an effective and scalable paradigm for advancing MoE-based LLMs.

In this paper, we make following contributions:

\begin{itemize}
    \item We design \ours, a method that breaks the limitation of \emph{layer-local} routing in MoE architectures. By reusing experts across adjacent layers while adding only negligible router parameters, \ours significantly increases the flexibility of expert combinations.
    \item We propose a \emph{Progressive Scaling Routing} strategy in \ours, which gradually enlarges the candidate expert pool during training, thereby reducing language modeling loss and improving downstream task accuracy.
    \item We have conducted extensive experiments to demonstrate that~\ours consistently improves both language modeling ability and downstream task performance across different model sizes and architectures, establishing \ours as a practical design paradigm for parameter-efficient and scalable MoE-based LLMs.
\end{itemize}

\vspace{-1em}

\section{Related Works}

\paragraph{Mixture-of-Experts.}
The strength of large models lies in their vast parameter counts, but this also brings the challenge of high computational cost.
The Mixture-of-Experts (MoE) framework was introduced to decouple parameter size from per-token computation in large language models (LLMs) during both training and inference.
In MoE-based transformer architectures~\citep{vaswani2017attention}, sparse MoE blocks~\citep{shazeer2017outrageously,fedus2022switch,lepikhin2020gshard} replace the Feed-Forward Networks (FFNs), improving efficiency while preserving strong performance.
A notable example is Mixtral-of-Experts (Mixtral MoE)~\citep{jiang2024mixtral}, an open-source architecture that activates 2 experts from a pool of 8. 
Compared with dense models of similar computational cost, Mixtral delivers stronger performance across multiple downstream tasks.
More recently, DeepSeekMoE~\citep{dai2024deepseekmoe} adopts a fine-grained MoE design by dividing each FFN into smaller experts, enabling more flexible expert combinations without inflating the total parameter count.
The open-source community has continued to advance this trend toward finer-grained experts. For instance, the Qwen3 series~\citep{Yang2025Qwen3TR} employs 128 experts, while Kimi-K2~\citep{kimi-k2} scales to 384 experts. 
Both demonstrate strong performance across diverse domains, reinforcing the idea that richer expert combinations often translate into better results.
Overall, the success of these models highlights \textit{finer-grained} design as a reliable and promising direction for future MoE architectures.

\paragraph{Parameter Reusing in Transformers.}
The standard Transformer constructs token representations using a stack of $L$ distinct layers, each consisting of a self-attention mechanism and a feed-forward network (FFN).
Recent studies~\citep{dehghani2018universal,csordas2024moeut,bae2024relaxed} have explored reusing a shared set of weights across multiple layers, showing promising gains in parameter efficiency.
Universal Transformers~\citep{dehghani2018universal} replace the standard stack of unique layers with a single parameter-shared block that is applied recurrently, refining token representations in parallel. 
This combines the inductive bias of RNNs with the parallelization benefits of Transformers.
MoEUT~\citep{csordas2024moeut} extends this idea by integrating the Mixture-of-Experts (MoE) paradigm into the recurrent Universal Transformer architecture, addressing its parameter–compute scaling bottleneck. 
This method increases model capacity while maintaining computational efficiency, making parameter sharing feasible for large-scale language modeling.
Relaxed Recursive Transformers~\citep{bae2024relaxed} further relax the strict layer-tying constraint by introducing depth-wise low-rank adaptation (LoRA) modules, improving performance while keeping the overall model compact.
A related approach to parameter reuse in MoE blocks is WideNet~\citep{xue2022go_widenet}, which derives its strategy from the perspective of reducing total parameters by recurrently reusing the weights of FFNs and self-attention blocks across all Transformer layers.
Experiments on small-scale models for both CV and NLP tasks demonstrate its effectiveness, highlighting parameter sharing as a practical way to improve parameter efficiency.

\section{Method}

In this section, we first revisit the widely used TopK routing strategy for Mixture-of-Experts (MoE) models. 
We then introduce~\ours, which enlarges the candidate expert pool by reusing experts across adjacent layers. 
To further improve performance when increasing the number of routed experts, we propose a Progressive Scaling Routing (PSR) strategy for training. 
An overview of~\ours is shown in~\autoref{fig:ours}.

\subsection{Review of TopK Routing Mixture-of-Experts}

In a standard $L$-layer transformer-based Mixture-of-Experts (MoE) architecture, the Feed-Forward Network (FFN) blocks are replaced with MoE blocks, each comprising $N$ experts and a router. 
The candidate expert pool of layer-$l$, $\mathcal{E}^l = \{\mathrm{E}_1, \mathrm{E}_2, \dots, \mathrm{E}_N\}$, consists of $N$ experts, each instantiated as an independent FFN.
The router is responsible for assigning each input token to a subset of experts. 
Specifically, the router utilizes the gating network, which is parameterized by trainable weights, computes the probability distribution for the given input, then selects the corresponding experts according to its routing strategy. 
In TopK routing MoE, the output of the MoE block in $l$-th layer is computed as follows:
\begin{align}
  \mathbf{h}' &= 
  \sum_{i=1}^{N} \bigl( \mathrm{g}_i\, \mathrm{E}_i (\mathbf{h}) \bigr) 
  \label{eq:top_k} \\[6pt]
  \mathrm{g}_i &= 
  \begin{cases}
      s_{i}, & s_{i} \in \mathrm{TopK}\left(\{s_j \mid 1 \leq j \leq N\}\right), \\
      0, & \text{otherwise},
  \end{cases}
  \label{eq:topk} \\[6pt]
  \mathbf{s} &= \mathrm{Softmax}(\mathbf{W} \cdot \mathbf{h})
  \label{eq:softmax}
\end{align}
where $\mathbf{W} \in \mathbb{R}^{N \times d} $ is the weight of the gating network in the router, and $g_i$ is the gating score for expert-$i$.
For brevity, we omit the self-attention and layer normalization in the above formulations.

\subsection{Expanding Candidate Expert Pool}

To overcome the limitation of the \textit{layer-local} routing mechanism, we expand the candidate expert pool by allowing the router to select from experts in \textit{grouped} adjacent layers.
Consider an $L$-layer MoE. Let $r$ denote the expert reuse frequency across layers, and let $\mathcal{E}^i$ represent the candidate expert pool of the $i$-th layer.
In~\ours, the grouped candidate expert pool for layer $l$ is defined as:

\vspace{-0.5em}
\begin{align}
    \mathcal{U} & := \bigcup_{i \in G} \mathcal{E}^i, \ \ \ \ G = \{\bigl\lfloor l / r \bigr\rfloor+k \mid 1 \leq k \leq r\}
\end{align}

Here, group $G$ is formed by $r$ consecutive layers starting from the $\lfloor l / r \bigr\rfloor$-th layer.
In this way, the candidate expert pool of each layer becomes $r$ times larger than in the vanilla setting.
The computation of each layer’s MoE block is then formulated as:
\begin{align}
  \mathbf{h}' &= 
  \sum_{i=1}^{rN} \bigl( \mathrm{g}_i\, \mathrm{U}_i (\mathbf{h}) \bigr)
  \label{eq:rex_moe}
\end{align}
where $\mathrm{U}_i \in \mathcal{U}$ is the $i$-th expert in the expanded pool.
By increasing $r$, the enlarged the candidate pool enables more diverse expert combinations and yielding performance gains.

\paragraph{Discussion.}
A specific router configuration can restrict routing to only local experts, in which case the model reduces to the vanilla MoE.
This guarantees that our method always matches the baseline performance under the most constrained setting.
When expert reuse is enabled, however, each MoE block includes a larger set of experts, which allows for more diverse expert combinations but can also lead to imbalanced routing patterns.
Consequently, load imbalance becomes a critical bottleneck that not only limits generalization but also introduces challenges during training.

\subsection{Progressive Scaling Routing Strategy}\label{sec:psr}

Another key component of~\ours is the Progressive Scaling Routing (PSR) strategy, which gradually increases the number of candidate experts during training.
When reusing experts from $r$ layers in a TopK MoE with $N$ experts per layer, each router can access up to $rN$ candidates.
Instead of training the router to select from all $rN$ candidates from the start, we adopt a progressive scheme: the number of available candidates begins at $N$ and is linearly expanded over the course of training.
At iteration $t$, the candidate expert pool size $N_t$ is defined as:

\begin{align}
    \mathrm{N}_t &= 
  \begin{cases}
      N, & t \leq t_s, \\
      \bigl \lfloor (1 + \frac{(r-1)(t-t_s)}{t_e-t_s}) N \bigr \rfloor, & t_s < t \leq t_e,
      \\
      rN, & t > t_e,
  \end{cases}
\end{align}

where $t_s$ and $t_e$ specify the start and end iterations of the scaling schedule, respectively.
At each iteration, we randomly mask $(rN - N_t)$ experts by setting their gating scores to zero before applying the TopK selection for each token.
This design follows the principle of curriculum learning, allowing the model to gradually learn richer and more diverse expert representations.

\section{Experiments}

\subsection{Experimental Setup}\label{sec:exp_setup}

\paragraph{Training environment.}
All models are trained with Megatron-LM~\citep{megatron-lm}, an open-source framework for large-scale language model training. 
We modified the \texttt{MoE Block} and \texttt{TopK Router} implementations to support cross-layer expert reuse and the Progressive Scaling Routing strategy during training. 
All models are pre-trained from scratch without instruction tuning, using the same hyperparameters across all runs. 
The sequence length is $4{,}096$ and the total batch size is 512, resulting in a global batch size of 2M tokens. 
For optimization, we use AdamW~\citep{loshchilov2017decoupled_adamw} with $\beta_1=0.9$, $\beta_2=0.95$, weight decay $0.1$, and a gradient clipping ratio of $1.0$. 
The learning rate is scheduled to start at $3 \times 10^{-4}$ and decay to $3 \times 10^{-5}$ following a cosine schedule. 
Further details are provided in Appendix~\ref{appsec:training_details}. 
All training jobs are conducted on 4 nodes, each equipped with 32$\times$ NVIDIA Hopper GPUs.

\paragraph{Model architecture.}  
\begin{table}[!t]
    \centering
    \small
    \setlength{\tabcolsep}{5pt}
        \caption{Base MoE architectures used in experiments.
        ``MoE-0.5BA0.07B'' denotes a MoE model with 0.5B total parameters and 0.07B active parameters per token.
        ``SE'' means ``Shared Experts''. This naming convention applies to all models.
    }
    \setlength{\tabcolsep}{3.2pt}
    \begin{tabularx}{\linewidth}{l|ccccc}
        \toprule
        \textbf{Model}             & \textbf{Hidden Size} & \textbf{Intermediate Size} & \textbf{\#Layers} & \makecell{\textbf{Heads}\\\textbf{(Q / KV)}} & \makecell{\textbf{\#Experts}\\\textbf{(Shared + Routed / Total)}} \\
        \midrule
        MoE-0.5BA0.07B    & 768       & 384       & 16       & 16 / 2          & 4 / 32     \\
        MoE-0.5BA0.07B-SE & 768       & 384       & 16       & 16 / 2          & 1 + 3 / 32 \\
        \midrule
        MoE-2.3BA0.3B     & 512       & 744       & 32       & 16 / 2          & 8 / 64     \\
        MoE-2.3BA0.3B-SE  & 512       & 744       & 32       & 16 / 2          & 2 + 6 / 64 \\
        \midrule
        MoE-7BA3B-SE      & 2048      & 1408      & 32       & 16 / 4          & 2 + 6 / 64 \\
        \bottomrule
    \end{tabularx}
    \label{tab:moe_arch}
    \vspace{-2em}
\end{table}
  
We adopt the widely used Mixture-of-Experts (MoE) transformer architecture with consistent dimensionality settings across all ablation studies.  
The only differences lie in the router parameters under different reuse configurations.  
The architectural configurations are summarized in~\autoref{tab:moe_arch}, where each model name specifies the number of activated and total parameters.  
The suffix ``-SE'' indicates that the architecture employs shared experts~\citep{dai2024deepseekmoe,rajbhandari2022deepspeed}, and~\oursshort models follow the same naming convention.  
In addition, ``-R\{$r$\}'' denotes that experts are reused across $r$ layers.

\paragraph{Training data.}
We use the sample-100BT partition\footnote{https://huggingface.co/datasets/HuggingFaceFW/fineweb-edu/viewer/sample-100BT} from fineweb-edu dataset~\citep{lozhkov2024fineweb-edu,penedo2024fineweb}.
The tokenizer is from LLaMA-2~\citep{touvron2023llama2}, with a vocabulary size of $32{,}000$. 
Since the vocabulary is relatively small, the LLaMA-2 tokenizer does not achieve a high compression ratio.
As a result, the processed 100B tokens cover around 87\% of the original text.
To ensure fair comparison, we fixed the data-parallel size and the shuffle seed, so that all experiments were trained on the same tokens in the same order, making the results directly comparable.

\paragraph{Evaluation metrics.} We use lm-evaluation-harness~\citep{lm-eval-harness} to evaluate performance on downstream tasks. Specifically, we report zero-shot accuracy on ARC-Easy (ARC-E) \& ARC-Challenge (ARC-C)~\citep{clark2018think_arc}, BoolQ~\citep{clark2019boolq}, HellaSwag~\citep{zellers2019hellaswag}, LAMBADA~\citep{paperno2016lambada}, LogiQA~\citep{liu2021logiqa}, OpenBookQA~\citep{mihaylov2018can_openbookqa}, PIQA~\citep{bisk2020piqa}, SciQ~\citep{welbl2017crowdsourcing_sciq}, SIQA~\citep{sap2019socialiqa_siqa} and WinoGrande~\citep{sakaguchi2021winogrande}.
For evaluation of the impact on inference speed after reusing experts from adjacent layers, we adapted~\ours to vLLM~\citep{kwon2023efficient_vllm} and report the throughput (tokens per second)
for prefill and decoding stages. 
Sampling is disabled in generation.

\subsection{Main Results}

\begin{table}[!t]
    \centering
    \small
    \setlength{\tabcolsep}{2.8pt}
    \caption{Comparison between~\ours and vanilla MoE models.
    All models are trained on 100B tokens. 
    Task abbreviations: \textbf{Hella.} = HellaSwag, \textbf{LAMB.} = LAMBADA, \textbf{Lg.QA} = LogiQA, \textbf{Op.QA} = OpenBookQA, \textbf{Wino.} = WinoGrande. 
    The best accuracy is highlighted in bold.}
    \begin{tabularx}{\linewidth}{l|ccccccccc|c}
        \toprule
        \textbf{Model}              & \textbf{ARC-E} & \textbf{Hella.} & \textbf{LAMB.} & \textbf{Lg.QA} & \textbf{Op.QA} & \textbf{PIQA}  & \textbf{SciQ}  & \textbf{SIQA}  & \textbf{Wino.} & \textbf{Avg.}$\uparrow$ \\
        \midrule
        MoE-0.5BA0.07B              & 50.67          & 38.38           & 32.37          & \textbf{28.42} & 31.00          & 65.29          & \textbf{71.20} & \textbf{38.84} & \textbf{53.04} & 45.47                   \\
        \oursshort-0.5BA0.07B-R2    & 52.31          & 39.06           & \textbf{33.75} & 25.65          & 32.80          & 65.78          & 71.10          & 38.33          & 51.22          & 45.56                   \\
        \oursshort-0.5BA0.07B-R4    & \textbf{53.91} & \textbf{39.46}  & 32.76          & 25.35          & \textbf{32.80} & \textbf{66.81} & 71.00          & 38.38          & 52.17          & \textbf{45.85}          \\
        \midrule
        MoE-0.5BA0.07B-SE           & 51.85          & 38.90           & 33.26          & 24.88          & 32.00          & 66.05          & 70.60          & \textbf{39.05} & 51.54          & 45.35                   \\
        \oursshort-0.5BA0.07B-SE-R2 & 52.06          & 39.28           & 32.43          & 26.57          & \textbf{35.00} & 66.54          & 71.80          & 37.41          & \textbf{51.93} & 45.89                   \\
        \oursshort-0.5BA0.07B-SE-R4 & \textbf{53.11} & \textbf{39.39}  & \textbf{34.00} & \textbf{28.88} & 33.40          & \textbf{67.46} & \textbf{71.90} & 38.69          & 50.36          & \textbf{46.35}          \\
        \midrule
        MoE-2.3BA0.3B               & 58.42          & 47.14           & 37.55          & 27.19          & 34.80          & 69.21          & 75.80          & 38.69          & \textbf{53.51} & 49.15                   \\
        \oursshort-2.3BA0.3B-R2     & \textbf{61.32} & 46.84           & 37.20          & \textbf{28.57} & 35.00          & 69.48          & \textbf{76.50} & \textbf{39.61}          & 52.33          & 49.65                   \\
        \oursshort-2.3BA0.3B-R4     & 60.94          & \textbf{47.96}  & \textbf{38.75} & 28.42          & \textbf{37.00} & \textbf{70.18} & 76.30          & 39.36 & 53.12          & \textbf{50.23}          \\
        \midrule
        MoE-2.3BA0.3B-SE               & 58.42          & \textbf{48.79}  & 38.13          & 25.35          & 37.00          & 69.53          & 75.00          & \textbf{40.28} & 52.17          & 49.41                   \\
        \oursshort-2.3BA0.3B-SE-R2  & \textbf{59.09} & 47.99           & 38.54          & 27.34          & 37.60          & 69.48          & 74.20          & 39.56          & 52.72          & 49.61                   \\
        \oursshort-2.3BA0.3B-SE-R4  & 58.71          & 48.59           & \textbf{39.01} & \textbf{28.26} & \textbf{39.00} & \textbf{70.67} & \textbf{76.10} & 39.66          & \textbf{52.80} & \textbf{50.31}          \\
        \bottomrule
    \end{tabularx}
    \vspace{-2em}
    \label{tab:main_results}
\end{table}

\subsubsection{Evaluation on Downstream Tasks}

\paragraph{Comparisons to vanilla MoEs.}
We report the accuracy on downstream benchmarks in~\autoref{tab:main_results}. 
The results show that the proposed~\ours models consistently outperform vanilla MoE baselines across different model scales and benchmark tasks. 
Overall, \ours achieves stable improvements in both R2 and R4 configurations, with R4 often delivering the highest average accuracy. 
For example, compared to the base MoE-2.3BA0.3B, the R4 model attains the best results on tasks such as HellaSwag, LAMBADA, OpenBookQA, PIQA, and SIQA, raising the average score to $50.23\%$, which clearly surpasses the baseline’s $49.15\%$. 
Similarly, under the ``SE'' setting, \ours-R4 outperforms the corresponding base MoE-2.3BA0.3B-SE. 
For smaller models in the MoE-0.5BA0.07B series, the advantage of \ours also remains consistent, where both R2 and R4 configurations yield notable gains in average accuracy over the baseline. 
More detailed task-wise accuracy trends during training can be found in~\autoref{app:fig:task_specific} and~\autoref{app:fig:task_specific_small} in the appendix. 
In summary, these results demonstrate that~\ours consistently improves performance across different model scales and architectures, particularly on reasoning and knowledge-intensive tasks, highlighting its robustness, scalability, and general effectiveness.

\paragraph{Comparisons to LLMs with equivalent effective parameters}
\vspace{-1em}
\begin{table}[h]
    \centering
    \small
    \setlength{\tabcolsep}{2pt}
    \caption{Comparisons between~\ours and open-source models.
        We report results for models with equivalent total or activated parameters on selected language understanding benchmarks. Our method achieves competitive or superior performance across tasks.
    }
    \begin{tabularx}{\linewidth}{l|c|c|ccccccc}
        \toprule
        \textbf{Model}                             & \makecell{\textbf{\#Act.}                                                                                                                           \\ \textbf{Params}} & \textbf{Data} & \textbf{ARC-E} & \textbf{Hella.} & \textbf{LAMB.} & \textbf{Lg.QA} & \textbf{PIQA} & \textbf{SciQ} & \textbf{Wino.} \\
        \midrule
        Llama2-7B~\citep{touvron2023llama2}        & 7B/7B                     & 2T & \textbf{76.4} & \textbf{78.6} & \textbf{73.9} & 30.7          & 78.1        & 93.7           & 69.3         \\
        MPT-7B-Base~\citep{MosaicML2023MTP}        & 7B/7B                     & 1T & 67.3          & 76.1          & 70.3          & -              & 79.9         & -              & 68.3         \\
        \midrule
        DeepSeekMoE-16B~\citep{dai2024deepseekmoe} & 3B/16B                    & 2T & 68.1          & 77.1          & -              & -              & \textbf{80.2} & -              & \textbf{70.2} \\
        LLaMA-MoE-8B~\citep{llama-moe}             & 3B/8B                     & -  & 60.2          & 70.8          & 66.6          & 30.6          & 77.5         & 84.2           & 63.6         \\
        OpenMoE-8B~\citep{xue2024openmoe}     & 2.1B/8B                   & 1T & 64.1          & 45.5          & -              & -              & 74.2         & -              & 60.3         \\
        \midrule
        \oursshort-7BA3B-SE-R3                     & 3B/7B                     & 1T & 75.7          & 69.0          & 63.9          & \textbf{33.2} & 75.0         & \textbf{94.2} & 65.9         \\
        \bottomrule
    \end{tabularx}
    \label{tab:main_results_os_models}
\end{table}

We compare~\ours with representative open-source dense and MoE models of similar total or activated parameter scales in~\autoref{tab:main_results_os_models}. 
For a fair comparison, we scale the training data of~\oursshort-7BA3B-SE-R3 to 1T tokens sampled from fineweb-edu. 
The model exhibits well-balanced performance, achieving highest results on LogiQA and SciQ, even when compared to Llama2-7B~\citep{touvron2023llama2}, which uses more activated parameters and is trained on a larger corpus. 
Meanwhile,~\ours remains highly competitive across the other benchmarks. 
These results demonstrate the effectiveness of~\ours as model size and training data increase, highlighting its scalability for high performance.

\subsubsection{Impact on Inference Speed}

\begin{figure}[h]
  \centering
  \begin{subfigure}[t]{0.5\textwidth}
    \centering
    \includegraphics[width=\linewidth]{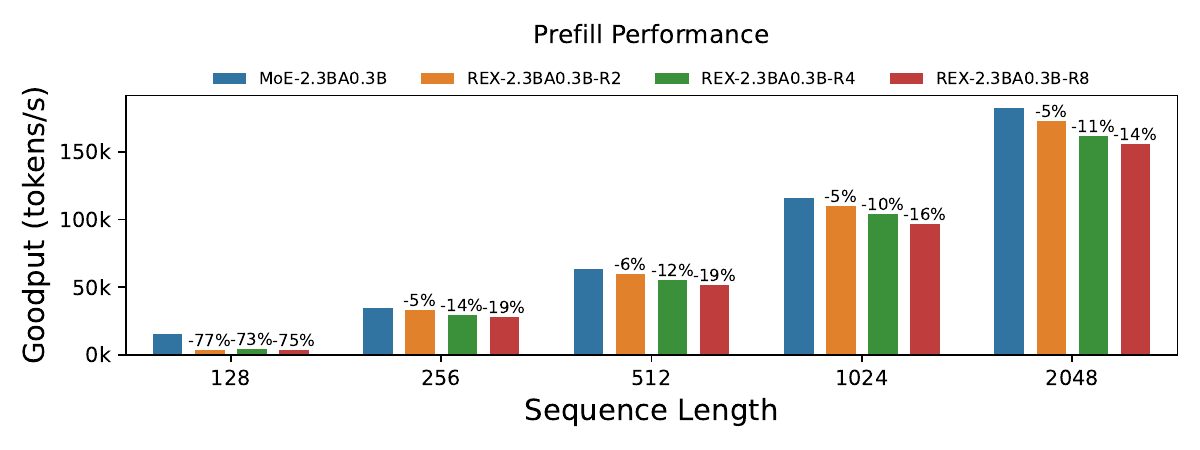}
    \caption{Prefill goodput under different sequence length.}
    \label{fig:inference_speed:a}
  \end{subfigure}\hfill
  \begin{subfigure}[t]{0.5\textwidth}
    \centering
    \includegraphics[width=\linewidth]{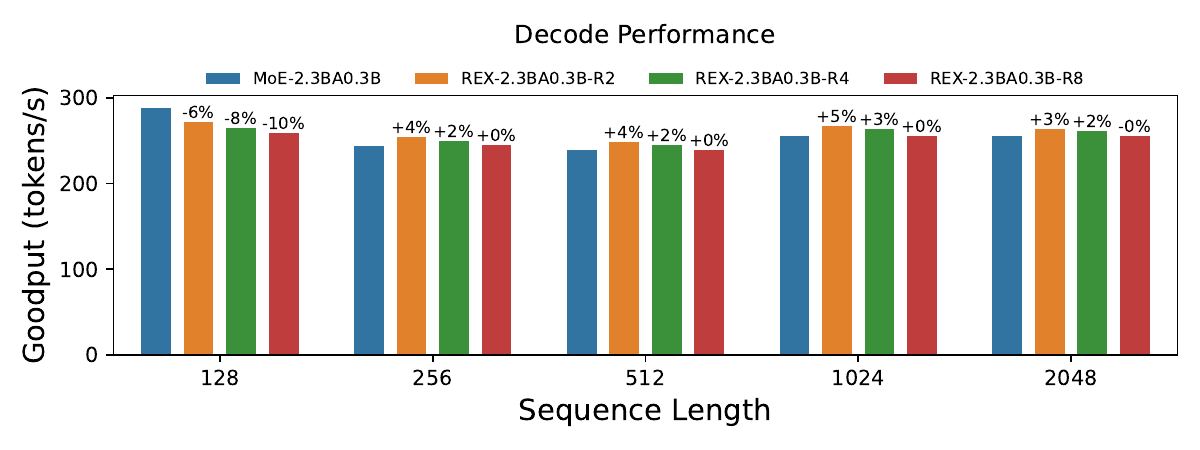}
    \caption{Decode goodput under different sequence length.}
    \label{fig:inference_speed:b}
  \end{subfigure}\hfill
\centering
    \caption{Comparison of prefill and decode goodput between base MoE and~\oursshort models.
    Numbers above the bars indicate the relative speedup over the base MoE.}

  \label{fig:inference_speed}
\end{figure}

We adapt~\ours to the vLLM inference engine~\citep{kwon2023efficient_vllm} to evaluate the impact of expert reuse on practical applications. 
We fix the output length at 128 tokens and vary the input length to assess both prefill and decoding performance across different sequence lengths. The detailed results are shown in~\autoref{fig:inference_speed}.
Although the computational overhead compared to vanilla MoE is negligible,~\ours introduces a larger number of experts into each MoE block, which increases I/O operations during the prefill stage. 
As a result, the inference speed of the reuse scheme experiences a noticeable decline. 
As shown in~\autoref{fig:inference_speed:a}, a larger candidate expert pool leads to slower prefill speed, with the performance degradation being more pronounced when the input length is relatively short.
Since the prefill stage usually accounts for only a small portion of the total time, the decoding stage is of greater practical importance. 
As illustrated in~\autoref{fig:inference_speed:b}, ~\ours achieves comparable performance across different sequence lengths in decoding stage.

\subsection{Ablation Studies}

\subsubsection{Effect of each component in~\ours}

\begin{wraptable}{r}{0.45\linewidth}
    \centering
    \vspace{-10pt}
    \setlength{\tabcolsep}{2.8pt}
    \caption{Average accuracy on benchmarks and PPL on WikiText.}
    \begin{tabularx}{\linewidth}{l|cc}
        \toprule
        \textbf{Model}             & \textbf{Avg. Acc}.$\uparrow$ & \textbf{PPL}$\downarrow$ \\
        \midrule
        MoE-2.3BA0.3B          & 49.15               & 21.19                      \\
        \ + Expert Reuse (4)  & 49.28               & 21.12                      \\
        \ + PSR           & \textbf{50.23}      & \textbf{20.73}             \\
        \bottomrule
    \end{tabularx}
    \label{tab:ablate_psr}
    \vspace{-10pt}
\end{wraptable}

To demonstrate the effectiveness of each component in~\ours, we provide comparative evaluations in ~\autoref{tab:ablate_psr}. 
We utilize the MoE-2.3BA0.3B model as our baseline and set the expert reuse frequency across layers as 4. 
In addition to using the same average accuracy over the benchmarks reported in~\autoref{tab:main_results}, we evaluate the validation perplexity (PPL) on WikiText~\citep{merity2016pointer_wikitext}.
We find that the simple expansion of the experts' pool results in only a marginal improvement. 
Specifically, a 0.13\% increase in averaged accuracy on downstream tasks and $0.07$ in PPL.
Furthermore, the incorporation of the PSR strategy yields a significant improvement in model performance by $1.05\%$ in the average accuracy and a drop in PPL of $0.46$.
Comprehensive benchmarks results of each task can be found in~\autoref{app:tab:ablation_full_results}.
These results demonstrate the effectiveness of the expert reuse and PSR strategy.

\subsubsection{Effect of Scaling Expert Reuse Group Size}

\begin{figure}[t]
  \centering
  \begin{subfigure}[t]{0.32\textwidth}
    \centering
    \includegraphics[width=\linewidth]{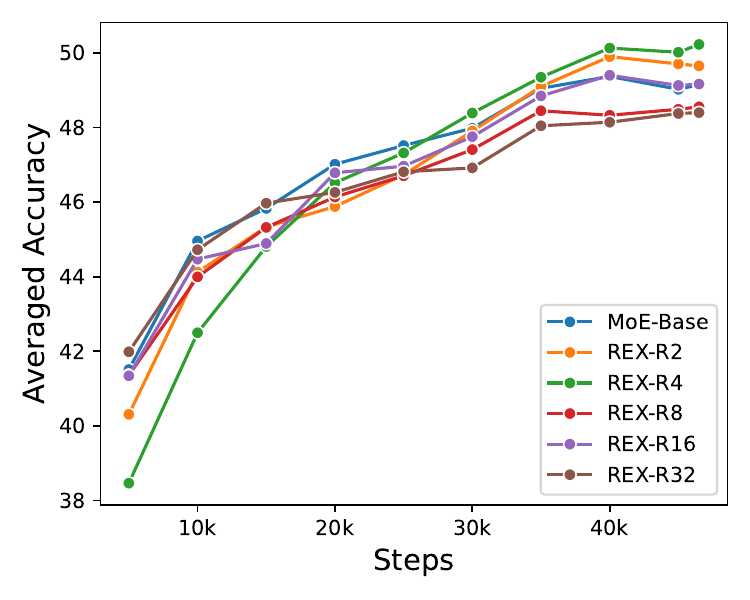}
    \caption{Average accuracy.}
    \label{fig:abalte_group_size:bmks}
  \end{subfigure}
  \begin{subfigure}[t]{0.32\textwidth}
    \centering
    \includegraphics[width=\linewidth]{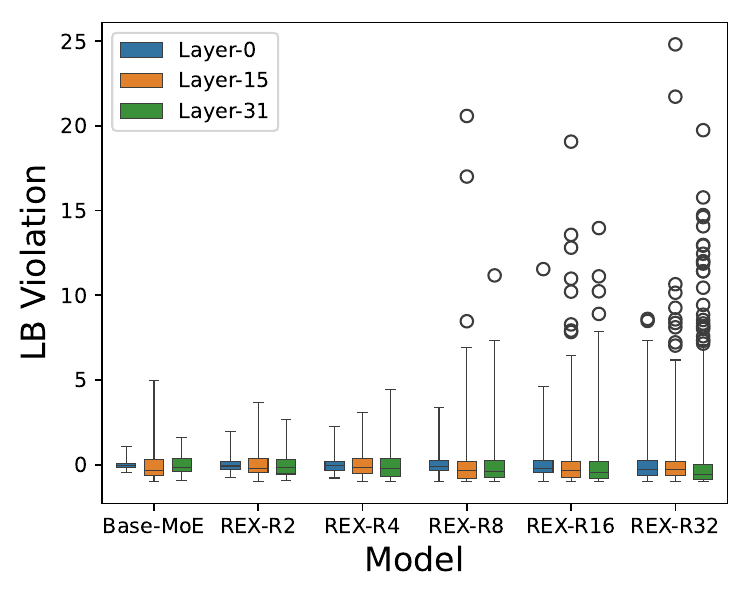}
    \caption{LB violation distribution.}
    \label{fig:abalte_group_size:vio}
  \end{subfigure}\hfill
  \begin{subfigure}[t]{0.32\textwidth}
    \centering
    \includegraphics[width=\linewidth]{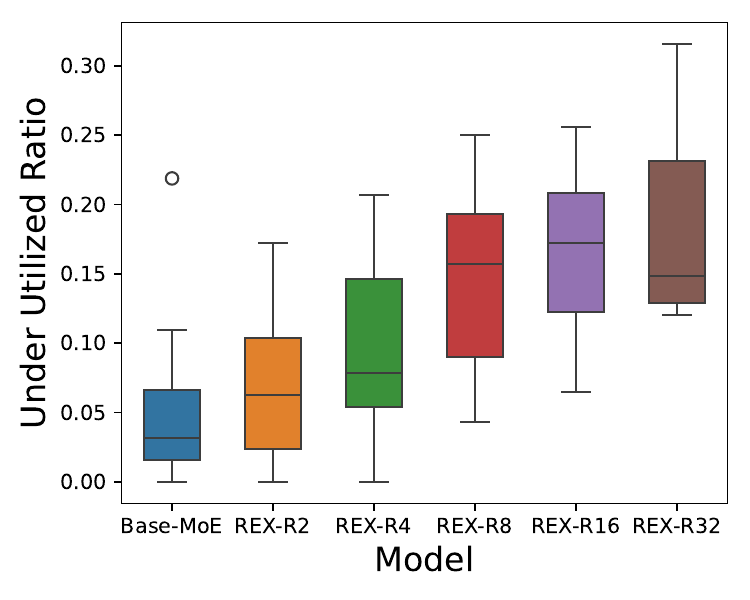}
    \caption{Ratio of under-utilized experts.}
    \label{fig:abalte_group_size:under}
  \end{subfigure}\hfill
  \caption{Average accuracy during training, distribution of load balance violations (degree of expert load imbalance), 
  and distribution of under-utilized experts ratios (larger values indicate more inactive experts) under different cross-layer expert reuse sizes.}
  \label{fig:abalte_group_size}
  \vspace{-1em}
\end{figure}

We investigate the impact of scaling the expert reuse group size in the 2.3B variant of~\ours, where the reuse frequency ranges from 2 to 32 layers. 
As presented in~\autoref{fig:abalte_group_size:bmks}, we illustrate the performance trends on downstream tasks during training for different configurations. 
In the early training phase, both \oursshort-R2 and \oursshort-R4 underperform the baseline MoE model; however, they eventually surpass it as training progresses, with larger reuse group sizes generally leading to better performance. 
In contrast, \oursshort-R16 and \oursshort-R32 initially match or even exceed the baseline but later fall behind in the later stages of training. 
More detailed evaluation results can be found in~\autoref{app:tab:ablation_group_size} in the appendix.
These results suggest that maintaining an appropriate balance in the number of reused layers is critical for sustaining high performance throughout training. 

To investigate the cause of the performance decline as the number of reused layers increases, we reserve a validation set from the C4 corpus\footnote{https://huggingface.co/datasets/allenai/c4} to evaluate load balance. 
Following the MaxVio metric in~\citep{wang2024auxiliary}, we adopt the Load Balance Violation (LBV) metric to quantify the degree of load imbalance in the MoE block. 
Specifically, the LBV of expert $i$ is computed as:
\begin{align}
    \mathrm{LBV}_i & = \frac{Load_i - \overline{Load_i}}{\overline{Load_i}}
\end{align}
where $Load_i$ denotes the number of tokens assigned to expert $i$, and $\overline{Load_i}$ is the average expert load. 
Under perfect balance, $\mathrm{LBV}_i$ equals $0$. 
As shown in~\autoref{fig:abalte_group_size:vio}, larger $r$ values lead to more significant deviations among outliers in the distribution of $LBV_i$, indicating that the model tends to activate only a few experts and suffers from a collapse phenomenon during training. 
In addition, we present the distribution of under-activated experts in~\autoref{fig:abalte_group_size:under}. 
We observe that as the candidate pool further expands, more experts remain barely activated throughout training. 
This imbalance in expert utilization explains why the performance of~\oursshort-R16 and~\oursshort-R32 is lower than baselines.

\subsubsection{Effect of Progressive Scaling Routing}

\begin{wrapfigure}{r}{0.5\linewidth}
  \centering
  \vspace{-1em}
  \begin{minipage}{\linewidth}
    \includegraphics[width=\linewidth]{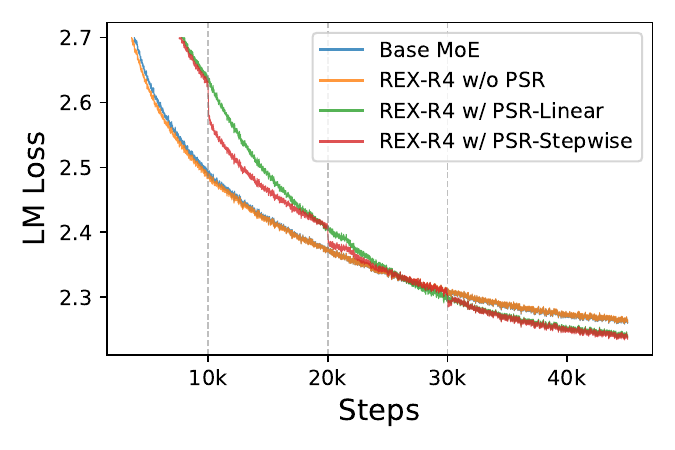}
    \vspace{-2em}
    \caption{Loss curves for different strategies.}
    \label{fig:ablate_psr}
  \end{minipage}
  \vspace{-1em}
\end{wrapfigure}

We investigate an alternative Progressive Scaling Routing (PSR) strategy to validate the optimal configuration adopted in our main experiments. 
Specifically, we introduce PSR-Stepwise, which keeps the number of candidate experts fixed over certain training intervals. 
In contrast, the strategy described in~\autoref{sec:psr} is referred to as PSR-Linear, as it provides a smoother and continuous expansion of the candidate expert pool. 
We use MoE-2.3BA0.3B as the baseline model and apply different PSR strategies to~\oursshort-R4. 
The corresponding training curves are shown in~\autoref{fig:abalte_group_size}. 
For both PSR-Stepwise and PSR-Linear, training starts with 64 candidate experts, and scaling begins at step $10$k. 
In PSR-Linear, the candidate pool is gradually increased to 256 by step $30$k. 
In PSR-Stepwise, the candidate pool is set to 128, 192, and 256 at steps $10$k, $20$k, and $30$k, respectively. 

We present the training loss curves of these models in~\autoref{fig:ablate_psr}. 
As shown in the figure, the loss curves of Base MoE and~\oursshort-R4, where PSR is not enabled, remain almost identical. 
When different PSR strategies are applied, model convergence is initially slowed. 
However, once the candidate pool begins to expand, the models achieve lower loss than those trained without progressive scaling. 
Notably, PSR-Stepwise accelerates loss reduction during the mid-training phase. 
\begin{wraptable}{r}{0.5\linewidth}
    \centering
    \caption{Average accuracy on benchmarks and PPL on WikiText.}
    \setlength{\tabcolsep}{3.2pt}
    \begin{tabularx}{\linewidth}{l|cc}
        \toprule
        \textbf{Model}          & \textbf{Avg. Acc.}$\uparrow$ & \textbf{PPL}$\downarrow$ \\
        \midrule
        \oursshort w/o PSR  & 49.28               & 21.12                      \\
        \oursshort w/ \ \ PSR-Stepwise & 49.59               & 20.76                      \\
        \oursshort w/ \ \ PSR-Linear   & \textbf{50.23}      & \textbf{20.73}             \\
        \bottomrule
    \end{tabularx}
    \label{tab:ablate_psr_exp}
    \vspace{-10pt}
\end{wraptable}
As summarized in~\autoref{tab:ablate_psr_exp}, both PSR strategies deliver clear performance improvements at final convergence, with detailed results provided in~\autoref{app:tab:ablation_full_results} in the appendix. 
Nevertheless, the final loss is comparable between the two strategies, while PSR-Linear achieves stronger overall performance on downstream tasks. 
Therefore, we adopt PSR-Linear to train all models. 

\subsection{Qualitative Analysis}

\begin{figure}[h]
  \centering
  \includegraphics[width=1\linewidth]{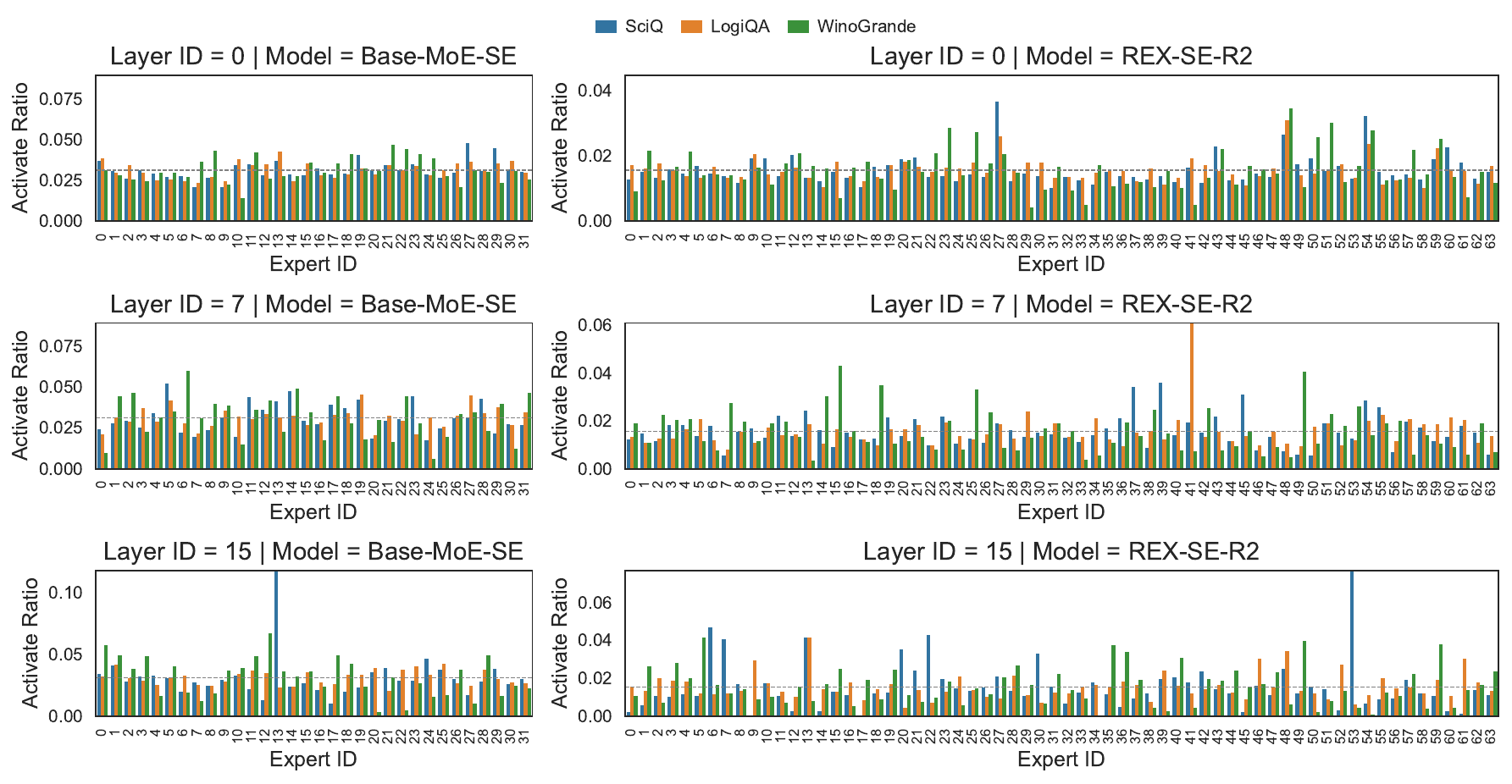}
  \caption{\textbf{Activate ratio of MoE-SE and~\oursshort-SE-R4 across layers in different tasks.}
  The gray dashed lines indicate uniform distribution. \ours shows stronger ability in task specialization.}
  \label{fig:act_ratio}
\end{figure}

In~\autoref{fig:act_ratio}, we present the expert activation ratios of Base-MoE-SE and~\oursshort-SE-R2 across layers 0, 7, and 15 on the SciQ, LogiQA, and WinoGrande tasks.
For Base-MoE-SE, the distribution of activated experts remains relatively uniform, with only minor variation across tasks.
In contrast, \oursshort-SE-R2 exhibits clear task-specific specialization.
For instance, certain experts (e.g., Experts 25 and 49) are activated far more frequently for WinoGrande than for the other tasks, especially in Layers 7 and 15.
Similar trends are observed on other tasks, as shown in~\autoref{app:fig:act_ratio_3} and~\autoref{app:fig:act_ratio_4} in the appendix.
These results suggest that the expanded expert pool of~\oursshort-SE-R2 allows for more effective task-specific allocation, encouraging the emergence of specialized experts and producing an ensemble-like effect in multi-task scenarios.

\section{Conclusion}

In this work, we present~\ours, a novel MoE design paradigm that overcomes the limitation of \textit{layer-local} routing. 
By allowing routers to reuse experts across grouped adjacent layers,~\ours decouples expert dimensionality from per-layer budgets and substantially enlarges the candidate expert pool with only negligible router overhead. 
Combined with the Progressive Scaling Routing strategy, it further enhances training stability and performance. 
Extensive experiments across diverse architectures and model scales show that~\ours consistently improves language modeling perplexity, downstream task accuracy, and the ability to learn task-specialized experts. 
Overall, these results establish~\ours as a parameter-efficient and practically scalable paradigm for designing MoE-based LLMs.

\newpage

\section*{Acknowledgement}

This work is partially funded by Research Council of Finland (grant number 362729) and Business Finland (grant number 169/31/2024).

\section*{Reproducibility statement}

We provide sufficient details for reproducing our key experiments. 
Training configurations are described in~\autoref{sec:exp_setup} and~\autoref{appsec:training_details}, while the data processing pipeline is detailed in~\autoref{appsec:data_proc}. 

\bibliography{iclr2026_conference}
\bibliographystyle{iclr2026_conference}

\newpage
\appendix

\section{The Use of Large Language Models (LLMs)}
\vspace{-1em}
We acknowledge the use of Large Language Models (LLMs) to assist in writing and polishing this paper. 
Their role was limited to improving the clarity and readability of the manuscript; they were not involved in the design of the methodology or in the scientific analysis.

\section{Additional Experiments Details}
\vspace{-1em}
\subsection{Data Processing}\label{appsec:data_proc}

We use the sample-100BT partition\footnote{https://huggingface.co/datasets/HuggingFaceFW/fineweb-edu/viewer/sample-100BT}
 of fineweb-edu~\citep{lozhkov2024fineweb-edu} for our main experiments.
Each sample in the dataset is tokenized independently and then randomly concatenated into sequences of $4{,}096$ tokens, which are used for training.

\subsection{Hyper-parameters and Parallelism Configurations}\label{appsec:training_details}

We use the same hyper-parameters for all model training runs.  
The training sequence length is set to 4,096, and the global batch size is 512, resulting in a training batch size of 2M tokens.  
The base frequency for Rotary Positional Embedding (ROPE)~\citep{su2024roformer_rope} is $10{,}000$.  
For optimization, we use AdamW~\citep{loshchilov2017decoupled_adamw} with $\beta_1=0.9$, $\beta_2=0.95$, and a weight decay of $0.1$, gradient clip ratio is $1.0$.
We adopt a warmup–cosine-decay learning rate scheduler, with an initial learning rate of $3 \times 10^{-4}$ that decays to $3 \times 10^{-5}$ by the end of training. The number of warmup steps is fixed at 100 for all experiments.  
When the number of routed experts exceeds $8$, we enable Expert Parallelism (EP) with a parallelism size of $8$ to accelerate training. 
No other parallelism strategies, such as Tensor Parallelism (TP) or Pipeline Parallelism (PP), are used in these runs.
We globally fix the random seed to $42$.

\section{Additional Experimental Results}

\subsection{Full Evaluation Results for Different PSR Variants}
\begin{table}[h]
    \centering
    \small
    \setlength{\tabcolsep}{2.5pt}
    \caption{Comparisons between base MoE and variants of~\ours.}
    \begin{tabularx}{\linewidth}{l|ccccccccc|c}
        \toprule
        \textbf{Model}                & \textbf{ARC-E} & \textbf{Hella.} & \textbf{LAMB.} & \textbf{Lg.QA} & \textbf{Op.QA} & \textbf{PIQA} & \textbf{SciQ} & \textbf{SIQA} & \textbf{Wino.} & \textbf{Avg.}$\uparrow$ \\
        \midrule
        Base MoE                      & 58.42          & 47.14           & 37.55          & 27.19          & 34.80          & 69.21         & 75.80         & 38.69         & 53.51          & 49.15                   \\
        \oursshort-R4  w/o PSR        & 58.16          & 46.94           & 38.52          & 25.96          & 36.40          & 70.67         & 74.50         & \textbf{39.46}         & 52.88          & 49.28                   \\
        \oursshort-R4 w/ PSR-Stepwise & 60.65          & \textbf{48.25}           & 37.67          & 27.04          & 34.40          & \textbf{70.84}         & 74.60         & 39.10         & \textbf{53.75}          & 49.59                   \\
        \oursshort-R4 w/ PSR-Linear   & \textbf{60.94}          & 47.96           & \textbf{38.75}          & \textbf{28.42}          & \textbf{37.00}          & 70.18         & \textbf{76.30}         & 39.36         & 53.12          & \textbf{50.23}                   \\
        \bottomrule
    \end{tabularx}
    \label{app:tab:ablation_full_results}
\end{table}

Complete evaluation results for different PSR variants are provided in~\autoref{app:tab:ablation_full_results}, with the base model being MoE-2.3B-A0.3B.

\subsection{Full Evaluation Results for Different Reuse Sizes}
\begin{table}[h]
    \centering
    \small
    \setlength{\tabcolsep}{2.8pt}
    \caption{Comparisons between base MoE and~\ours with different reuse sizes.}
    \begin{tabularx}{0.86\linewidth}{l|ccccccccc|c}
        \toprule
        \textbf{Model} & \textbf{ARC-E} & \textbf{Hella.} & \textbf{LAMB.} & \textbf{Lg.QA} & \textbf{Op.QA} & \textbf{PIQA} & \textbf{SciQ} & \textbf{SIQA} & \textbf{Wino.} & \textbf{Avg.}$\uparrow$ \\
        \midrule
        \oursshort-R8  & 58.75          & 46.80           & 37.07          & 26.27          & 35.00          & 69.97         & 72.50         & 37.97         & 52.64          & 48.55                   \\
        \oursshort-R16 & 58.59          & 46.79           & 38.48          & 27.80          & 35.40          & 70.02         & 72.20         & 39.36         & 53.83          & 49.16                   \\
        \oursshort-R32 & 58.21          & 46.28           & 35.26          & 27.04          & 35.60          & 70.35         & 72.80         & 39.15         & 50.91          & 48.40                   \\
        \bottomrule
    \end{tabularx}
    \label{app:tab:ablation_group_size}
\end{table}

Complete evaluation results for different reuse sizes are provided in~\autoref{app:tab:ablation_group_size}, with the base model being MoE-2.3B-A0.3B.

\subsection{Evaluation on Top2 MoE}

\begin{table}[h]
    \centering
    \small
    \setlength{\tabcolsep}{5pt}
    \caption{Architecture of Top2 MoE model used in the additional experiments.}
    \begin{tabularx}{\linewidth}{l|ccccc}
        \toprule
        Model         & Hidden Size & Intermediate Size & \#Layers & \makecell{Heads         \\(Q / KV)} & \makecell{\#Experts\\(Shared + Routed / Total)} \\
        \midrule
        MoE-0.5BA0.13B & 768         & 1536              & 16       & 16 / 2          & 2 / 8 \\
        \bottomrule
    \end{tabularx}

    \label{app:tab:additional_top2_architecture}
\end{table}

\begin{table}[h]
    \centering
    \small
    \setlength{\tabcolsep}{2.8pt}
    \caption{Comparisons between base MoE and~\ours with different reuse sizes.}
    \begin{tabularx}{\linewidth}{l|ccccccccc|c}
        \toprule
        \textbf{Model}          & \textbf{ARC-E} & \textbf{Hella.} & \textbf{LAMB.} & \textbf{Lg.QA} & \textbf{Op.QA} & \textbf{PIQA} & \textbf{SciQ} & \textbf{SIQA} & \textbf{Wino.} & \textbf{Avg.}$\uparrow$ \\
        \midrule
        MoE-0.5BA0.13B           & 51.85          & 38.70           & 33.09          & 27.34          & 33.00          & 66.05         & 67.40         & 38.54         & 51.38          & 45.26                   \\
        \oursshort-0.5BA0.13B-R2 & 52.82          & 39.26           & 33.18          & 27.96          & 32.00          & 66.05         & 70.60         & 38.08         & 52.80          & 45.86                   \\
        \oursshort-0.5BA0.13B-R4 & 51.94          & 39.34           & 32.25          & 27.04          & 32.80          & 65.56         & 70.60         & 38.69         & 50.51          & 45.41                   \\
        \bottomrule
    \end{tabularx}
    \label{app:tab:additional_top2_results}
\end{table}

We further apply~\oursshort to a Top2 MoE, with its architecture detailed in~\autoref{app:tab:additional_top2_architecture}.
The corresponding evaluation results are reported in~\autoref{app:tab:additional_top2_results}.

\subsection{Task-Wise Accuracy}

\begin{figure}[h]
    \centering
    \includegraphics[width=1\linewidth]{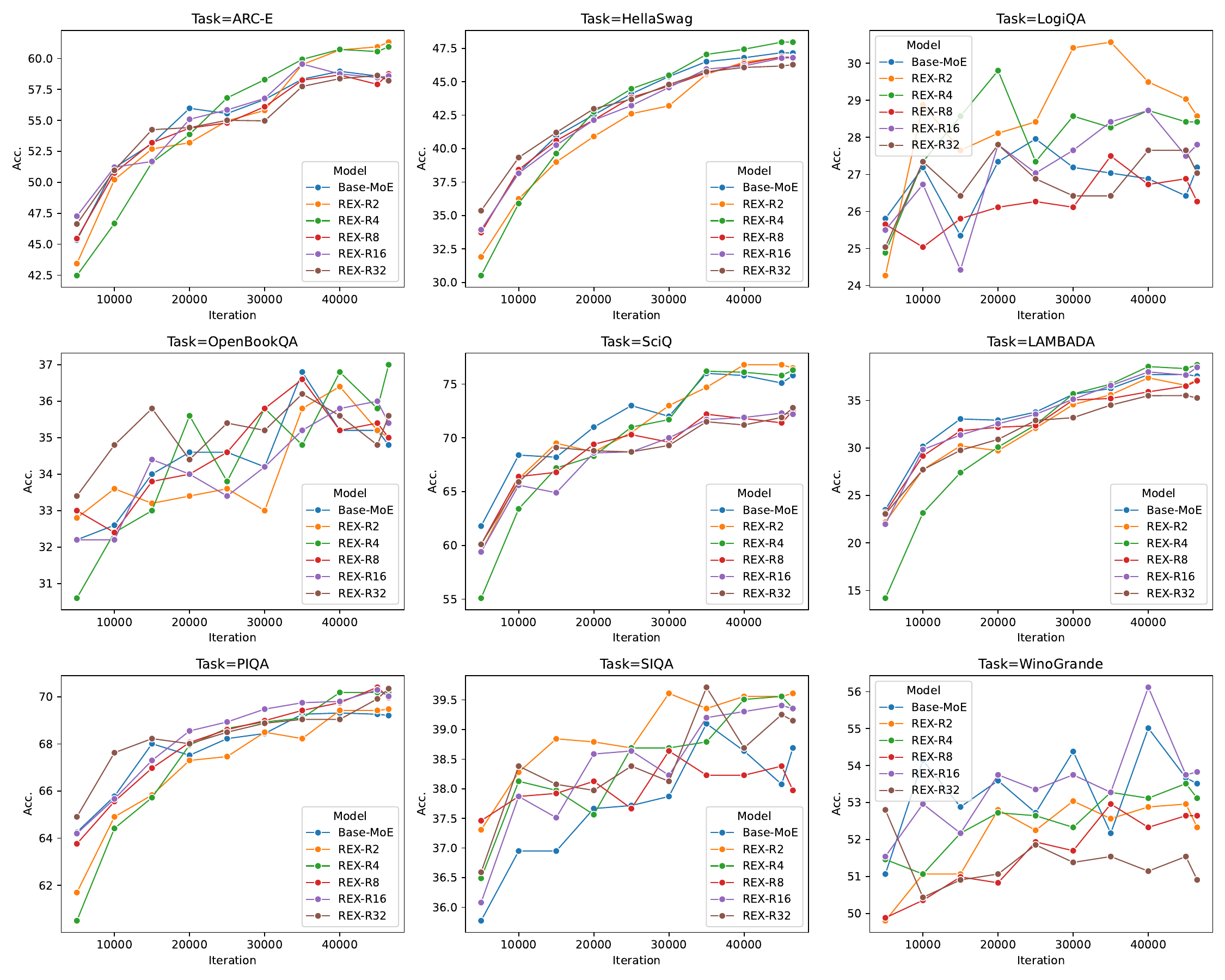}
    \caption{\textbf{Task-wise accuracy change as training progresses.} Base-MoE is MoE-2.3BA0.3B.}
    \label{app:fig:task_specific}
\end{figure}

\begin{figure}[ht]
    \centering
    \includegraphics[width=1\linewidth]{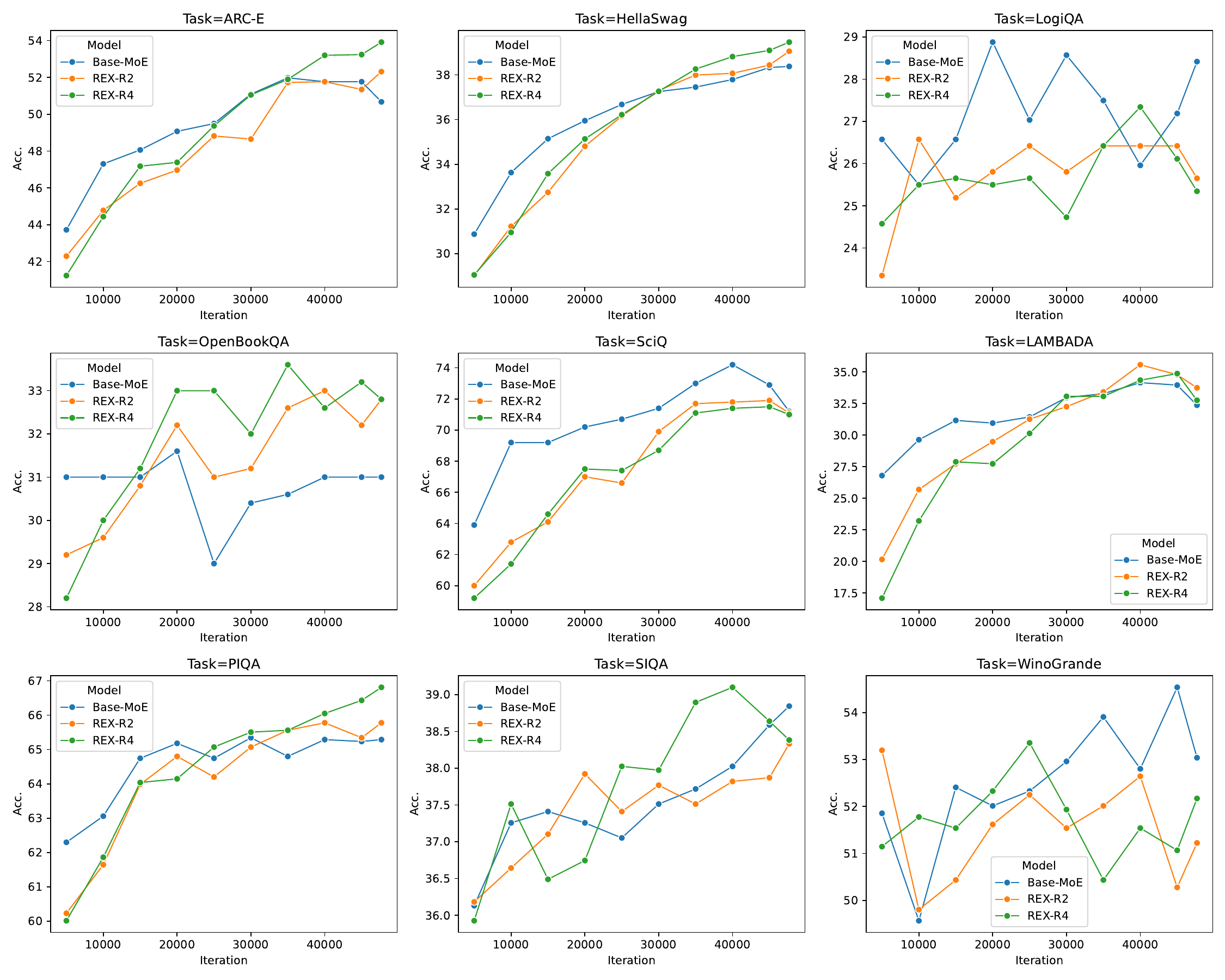}
    \caption{\textbf{Task-wise accuracy change as training progresses.} Base-MoE is MoE-0.5BA0.1B.}
    \label{app:fig:task_specific_small}
\end{figure}

\newpage
\subsection{Task-wise Experts Selection Visualization}

\begin{figure}[h]
  \centering
  \centering
  \includegraphics[width=1\linewidth]{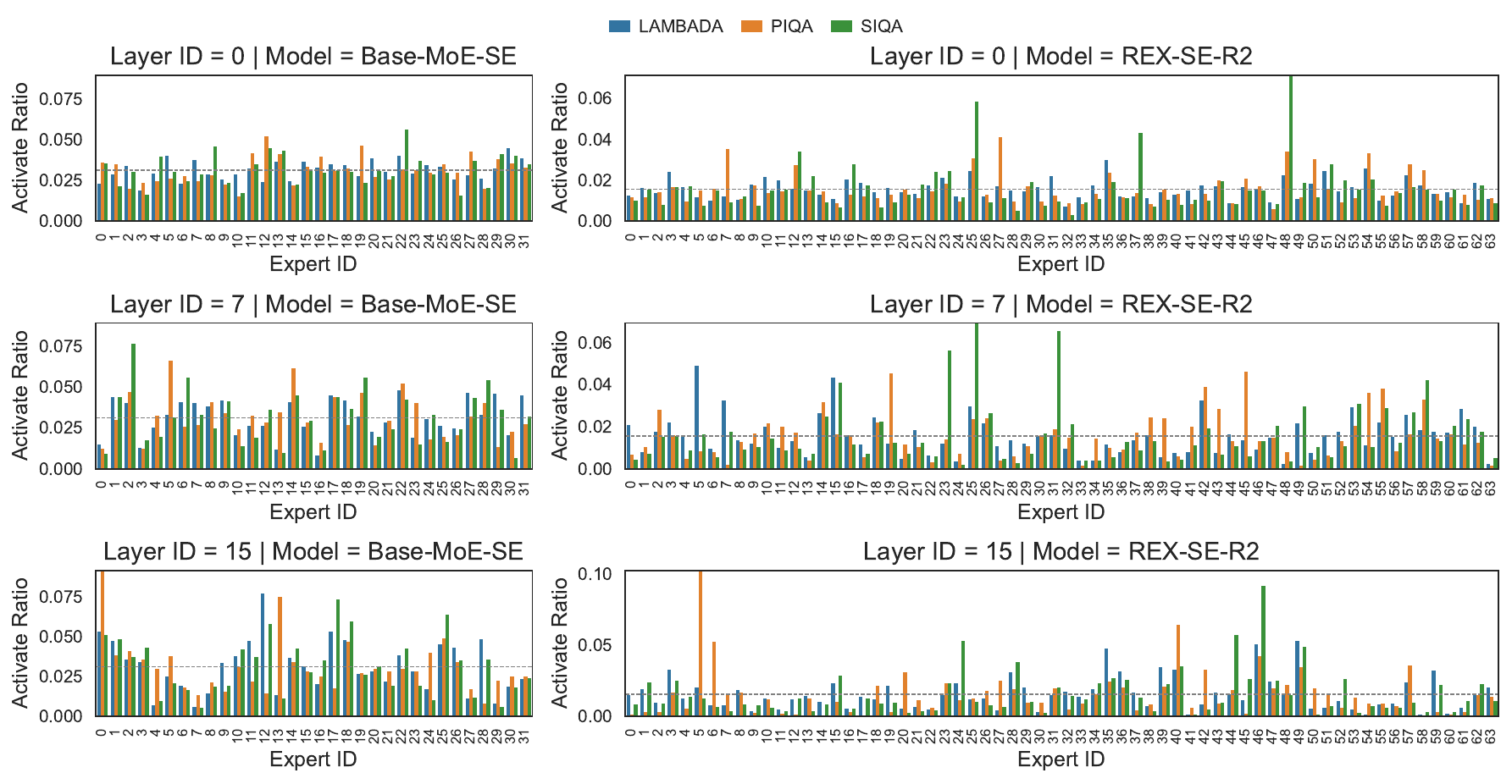}
  \caption{\textbf{Activate ratio of MoE-SE and~\oursshort-SE-R4 across layers in different tasks.}
  The gray dashed lines indicate uniform distribution.}
  \label{app:fig:act_ratio_3}
\end{figure}

\begin{figure}[h]
  \centering
  \centering
  \includegraphics[width=1\linewidth]{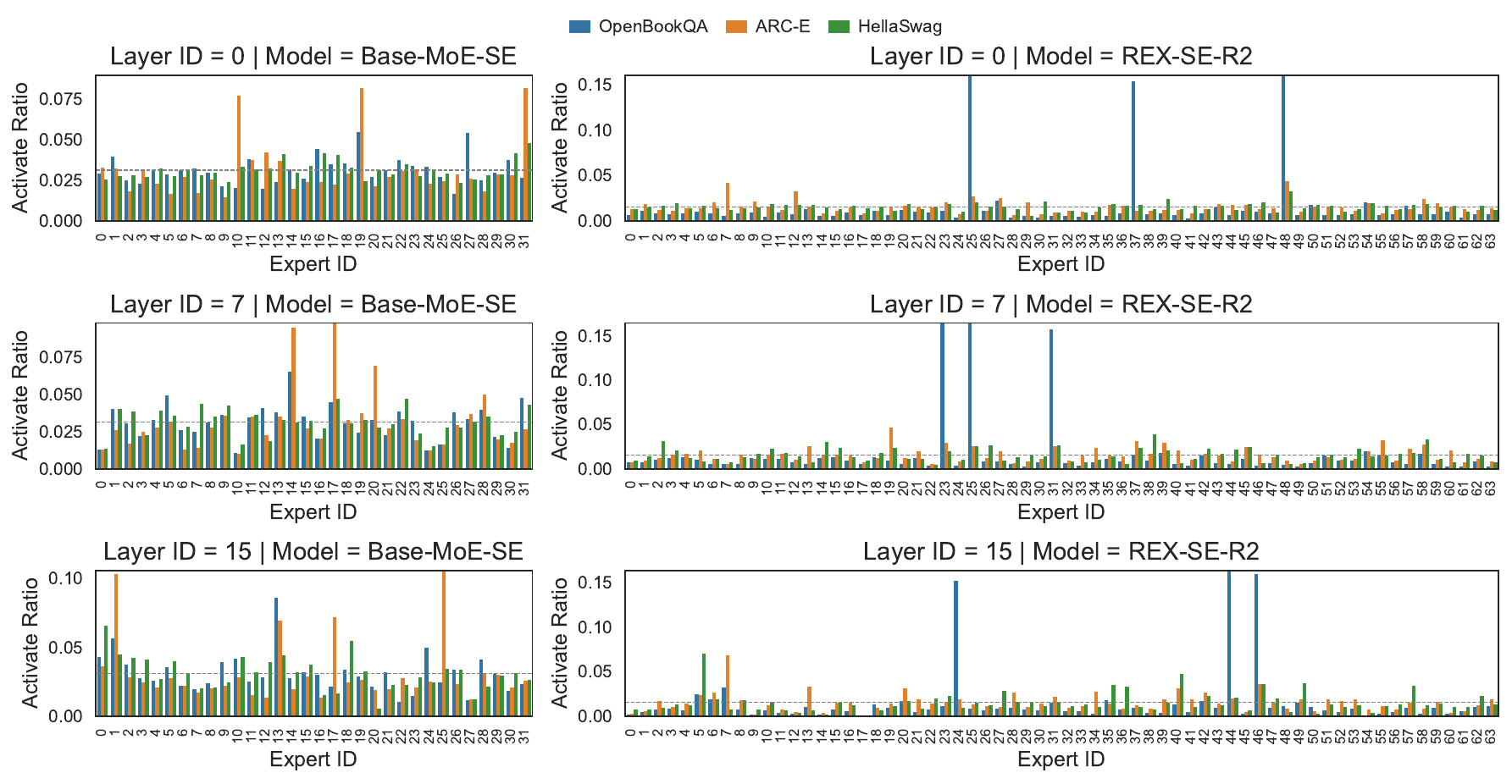}
  \caption{\textbf{Activate ratio of MoE-SE and~\oursshort-SE-R4 across layers in different tasks.}
  The gray dashed lines indicate uniform distribution.}
  \label{app:fig:act_ratio_4}
\end{figure}

\end{document}